\documentclass[10pt,twocolumn,letterpaper]{article}

\usepackage[pagebackref=true,breaklinks=true,letterpaper=true,colorlinks,bookmarks=false]{hyperref}
\usepackage[table,xcdraw]{xcolor}
\usepackage{wacv}
\usepackage{times}
\usepackage{epsfig}
\usepackage{graphicx}
\usepackage{amsmath}
\usepackage{amssymb}
\usepackage{siunitx}
\usepackage{array}

\usepackage{authblk}

\usepackage{adjustbox}
\usepackage{caption}
\usepackage{balance}

\usepackage{enumitem}

\newcommand{\floor}[1]{\lfloor #1 \rfloor}

\makeatletter
\def\BState{\State\hskip-\ALG@thistlm}
\makeatother

\wacvfinalcopy 

\def\wacvPaperID{348} 

\ifwacvfinal\fi
\begin{document}

\title{DeepWheat: Estimating Phenotypic Traits \\from Crop Images with Deep Learning}


\author[1]{\vspace{-6pt}
Shubhra Aich}
\author[2]{Anique Josuttes}
\author[1]{Ilya Ovsyannikov}
\author[2]{Keegan Strueby}
\author[1]{Imran Ahmed}
\author[2]{\\Hema Sudhakar Duddu}
\author[2,3]{Curtis Pozniak}
\author[2]{Steve Shirtliffe}
\author[1]{Ian Stavness\thanks{Corresponding Author: ian.stavness@usask.ca}}

\affil[1]{Dept. Computer Science, $^2$Dept. Plant Sciences, $^3$Crop Dev. Centre, Univ. Saskatchewan, Canada}



\maketitle

\ifwacvfinal\thispagestyle{empty}\fi

\begin{abstract}
In this paper, we investigate estimating emergence and biomass traits from color images and elevation maps of wheat field plots. We employ a state-of-the-art deconvolutional network for segmentation and convolutional architectures, with residual and Inception-like layers, to estimate traits via high dimensional nonlinear regression. Evaluation was performed on two different species of wheat, grown in field plots for an experimental plant breeding study. Our framework achieves satisfactory performance with mean and standard deviation of absolute difference of $1.05$ and $1.40$ counts for emergence and $1.45$ and $2.05$ for biomass estimation. Our results for counting wheat plants from field images are better than the accuracy reported for the similar, but arguably less difficult, task of counting leaves from indoor images of rosette plants. Our results for biomass estimation, even with a very small dataset, improve upon all previously proposed approaches in the literature.
\end{abstract}


\section{Introduction}

Measuring the phenotypic traits of crops, which are the differences in plant characteristics caused by the interaction of the plant's genetics and the environment, is important in plant breeding research as it allows the breeders to select crop varieties with desirable physical characteristics, such as high yield, resistance to stress, and ability to be easily harvested. Traditionally, phenotypic measurements are made manually in the field, which is both labor intensive and potentially inaccurate due to substantial sub-sampling involved. To overcome these drawbacks, image-based automated phenotypic traits estimation is emerging as an important area of applied computer vision research with the goal of capturing more accurate information at a large scale for better crop production.

\begin{figure}[t!]
\centering
	\includegraphics[scale=0.212]{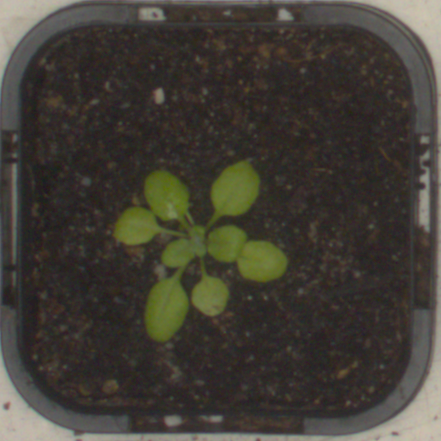}
	\includegraphics[scale=0.115]{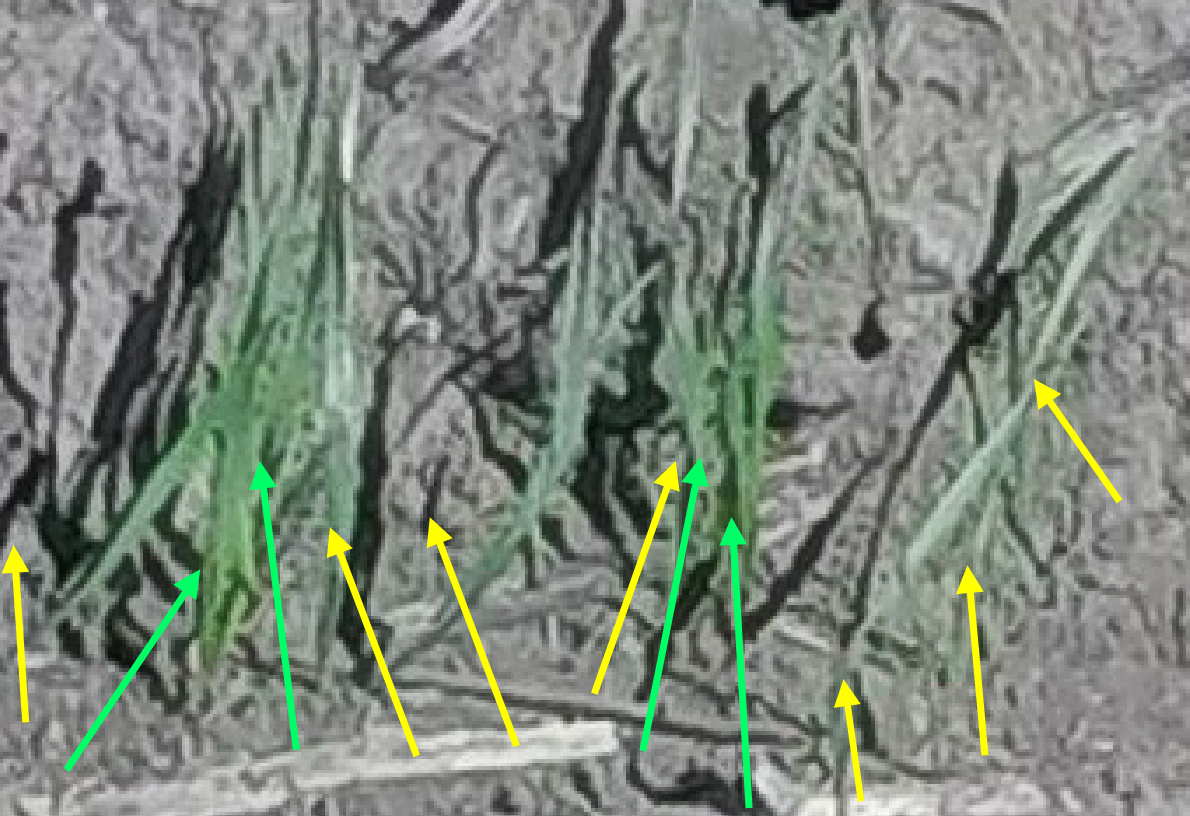}
    \caption{Eleven leaves in an image from the standard leaf counting dataset \cite{dataset-cvppp2017-03} (left) and eleven wheat plants in an outdoor image used for emergence counting in this paper. Counting plants from the right image is more challenging to due to variable number of leaves per plant and  occlusion.}
    \label{fig:emergence_difficult}
\end{figure}

In many crops, including wheat, emergence (the density of plants within the field) and biomass (the total mass of each plant) are important phenotypes. Emergence is important because a vigorous and uniform crop stand is needed to compete for moisture, nutrients, and sunlight. Plants that emerge late will have a lower yield than the early emerging ones due to the increase in competition for sunlight and essential nutrients \cite{lawles2012}. Determining biomass in different crop varieties is important because it is correlated with yield \cite{soriano2017}
and photosynthetic activity, and is an indicator of overall plant health \cite{dai2016}.
%
These phenotypes are labour intensive and destructive to measure manually: emergence typically requires physically touching plants in the field to determine which leaves belong to which plant, and biomass measurements are made by cutting out plants from the field and measuring their mass. Furthermore, these phenotypes are traditionally measured on only a small sub-sample of the experimental plot area, which can result in sampling error. The combination of high importance and high measurement difficulty makes these phenotypes good candidates for image-based phenotyping in any crop breeding programs.

Counting plants is related to the well-studied problem of counting leaves from plant images \cite{cvppp2015_winner, aich-cvppp2017}, but much more challenging. Wheat seeds are planted in close proximity, therefore, the plants grown from these seeds are highly occluded by each other in the image. To illustrate the level of difficulty, Figure \ref{fig:emergence_difficult} shows a sample image from the standard leaf counting dataset \cite{dataset-cvppp2017-03} and another image from the dataset we are using for wheat emergence counting. Both images have the same label: 11 leaves in the left image, and 11 wheat plants in the right image. In the left image, the number of leaves is unambiguous despite a few small leaves in the center, which is not the case for plant count in the right image. According to the plant science experts who generated the ground truth counts and who have experience counting plants in the field, while counting from the images, they looked at the stems as close to the ground as possible. When a stem seemed unreasonably thick, they presumed that there were more plants behind the visible ones. Plant bases indicated by the yellow arrows in the figure are easy to count. However, in regions denoted by the green arrows, it may look like there is one plant, based on the thickness of the plants, amount of leaves, and age of plants, the count of plants was estimated by the raters as more than one. Hence, both intuition and experience play a role in accurate emergence counting, making it a difficult image analysis task.

In this paper, we propose completely data-driven frameworks for emergence counting and biomass estimation.
We develop generalized architectures for phenotypic traits estimation blending the concepts of learning sparse structure via dense, multiscale representations \cite{inception-01} and residual or shortcut connections \cite{resnet}.
We train our models from scratch to keep our phenotypic estimation tasks independent of the other large-scale machine learning tasks pursued with very large models. For this reason, to efficiently train the data-hungry deep models with a few training samples, we also propose a novel data augmentation strategy based on randomized minimal region swapping of the superpixels in an image, which can be used to augment low to medium resolution images.
%
Also, we examine the quality of learning of the emergence counting architecture qualitatively by visualizing salient regions using the class activation mapping (CAM) \cite{cam-mit} approach. We find that the learned network features focus on image regions that are responsible for counting, notably the base of each leaf-cluster, and the dense regions of leaves, according to the plant breeding experts who provided the ground truth counts.


To the best of our knowledge, this is the first work on image-based phenotypic trait estimation of crops with deep learning. The name \textit{DeepWheat} refers to our overall system because of the first use of deep learning in this domain and since we have used the image dataset of two species of wheat for the evaluation. Although we evaluate our approach on wheat, our design allows the frameworks to be generalized to other types of crops with minimal additional manual intervention.


\begin{figure*}[t!]
	\centering
	\includegraphics[width=\textwidth]{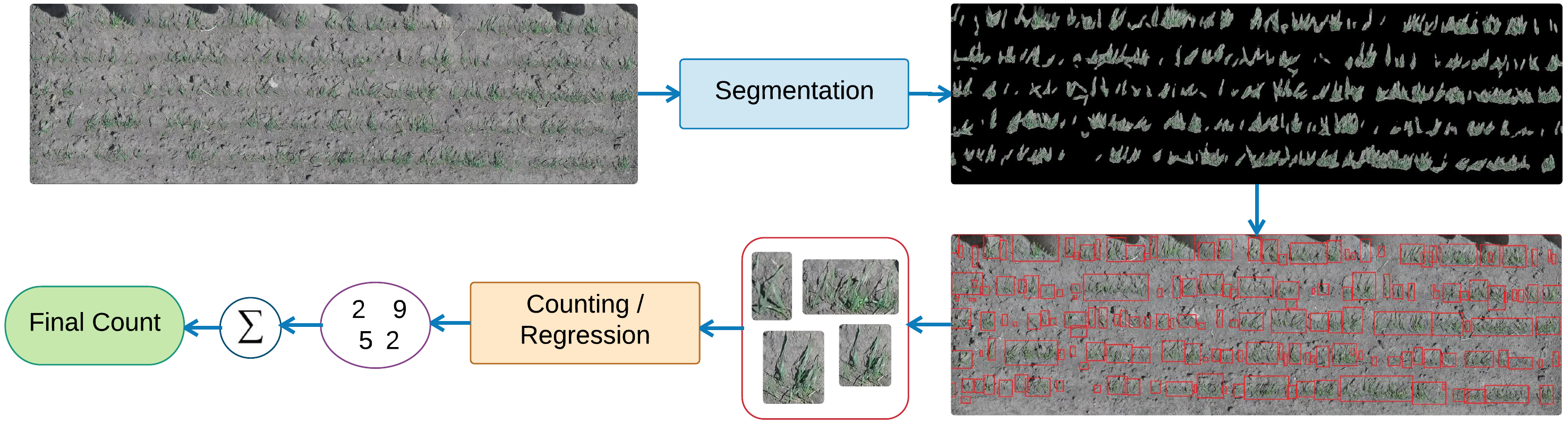}
    \caption{Workflow for emergence counting: 1) loosely segment the plant regions from RGB plot images with the segmentation module, 2) extract small patches containing plants via connected component analysis, 3) use counting module for individual counts on each patch, 4) sum all the patches to get the overall emergence count for a single plot.}
    \label{fig:workflow_emergence}
\end{figure*}

\begin{figure*}[]
	\centering
	\includegraphics[width=0.8\textwidth]{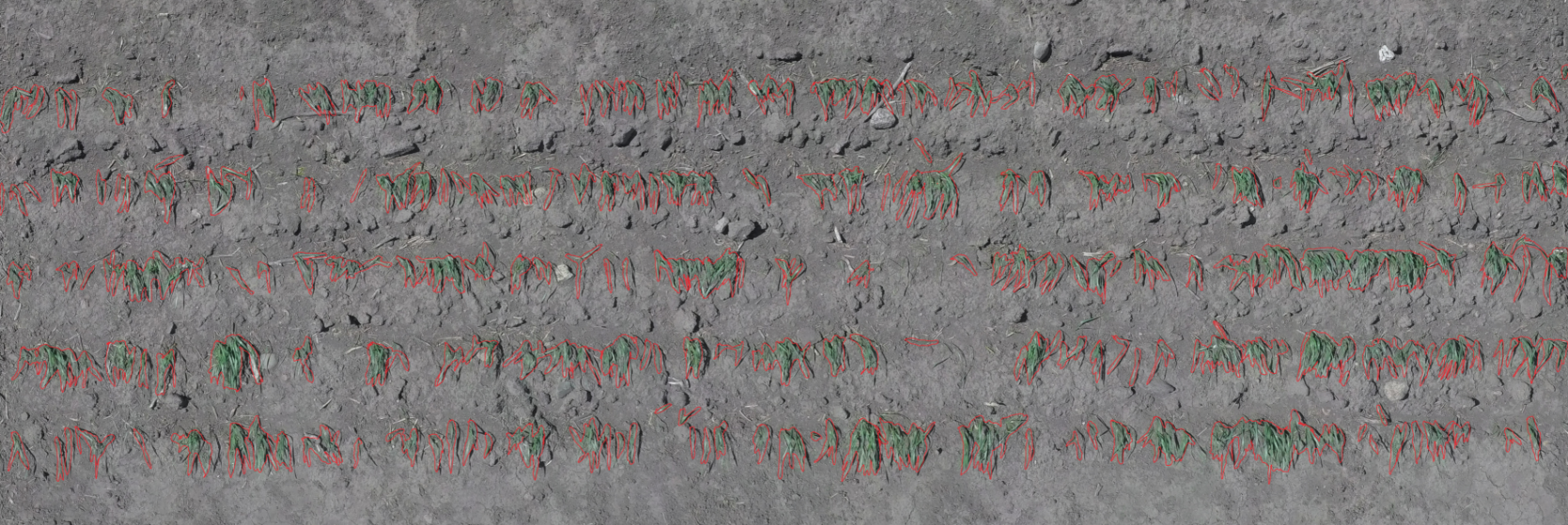}
    \caption{Manual ground-truth generated for relaxed segmentation of plants showing manually drawn contours around plant regions (red). Later, contours are filled with simple morphological hole-filling to create the binary segmentation mask.}
    \label{fig:gt}
\end{figure*}

\section{Related Work}
\label{sec:rw}

Despite the significance of emergence and biomass in crop breeding, little computer vision research has been done on the automated estimation of these traits from images. Leaf counting has been studied in more detail due to a standardized dataset of rosette plants and previous computer vision competitions \cite{lcc-2017}. Recent approaches to leaf counting have employed convolutional neural networks to count by regression \cite{aich-cvppp2017}. We adopt a similar approach in this study to evaluate if it extends to much more difficult phenotyping tasks such as plant and biomass counting from field images.

A few studies have looked at plant density estimation in maize~\cite{steward2001, steward2004, steward2005} and wheat~\cite{wheat-ann-2017, jin2017} from RGB images. All of these previous methods employ a traditional image processing pipeline that requires hand-tuned parameters tailored to the specific crop of interest.
In the wheat studies, the plant counting algorithm depends on the accurate segmentation of leaves, followed by extracting regional properties of the leaves as features, and then training a simple artificial neural network (ANN)~\cite{wheat-ann-2017} or a support vector machine (SVM)~\cite{jin2017}. In both papers, the initial segmentation of the plant foreground from the soil background is accomplished with simple naive approaches: Otsu thresholding on the ``b" channel of Lab image or a predefined RGB transformation channel $(2G-2B-2.4R)$. However, simple threshold-based segmentations are not robust to variable illumination in different field environments. Indeed, these segmentation approaches are found to give very poor results for the images used in our study and are therefore not useful benchmarks for comparison.


A number of previous studies have attempted to estimate biomass, but most have done so from field-based measurements and are therefore not applicable to image datasets. A few studies have used aerial images as a basis for biomass estimation. In \cite{schirrmann2016}, naive linear regression models are fitted on plant height and plant coverage in aerial images. In \cite{reddersen2014}, different linear and nonlinear combinations of height measured with an ultrasonic sensor, leaf area index measured with a plant canopy sensor, and vegetation indices from canopy reflectance obtained using a portable spectrometer are used as the predictors and biomass is used as the response of the multiple linear regression model. The product of leaf area index and dry matter content per leaf area is regarded as the estimation of above-ground biomass (AGB) in \cite{radiative2017}. The authors also provide a comparison against the models developed using exponential regression, partial least square regression and simple artificial neural networks. In \cite{laurin2016}, AGB was estimated from height information obtained from the Digital Terrain Model (DTM) derived from LiDAR data. For each plot, simple statistical measures of height, such as mean, quadratic mean, standard deviation, skewness, kurtosis, and percentile of height along with height bins at fixed intervals, are used as the predictors for regression modeling. A similar approach is taken in \cite{laurin2014} with additional vegetation indices extracted from hyperspectral data. In terms of the list of predictor variables, the approach in \cite{bendig2015} can be considered an extended version of the other two \cite{laurin2014, laurin2016} with height information plus the vegetation indices based on both hyperspectral and unmanned aerial vehicle (UAV) images.

\section{Our Approach}
\label{sec:approach}
In this section, we describe the design of both emergence count and biomass estimation frameworks in detail. Although both traits are estimated by convolutional networks performing regression, the architectures and overall workflows are different.

\subsection{Emergence Counting}
Figure \ref{fig:workflow_emergence} depicts the overall computational procedure for counting crop emergence. First, we loosely segment the plant regions from the RGB plot images through the segmentation module described later. Next, we extract all the segmented patches from the whole image, as indicated by the red rectangles in Figure \ref{fig:workflow_emergence} and input each patch image to the counting module to get the individual emergence counts for each patch. Finally, we sum up all the predicted counts for a single plot image to get the overall prediction for emergence count for that particular plot. In this framework, both the segmentation and the counting modules comprise deep architectures which we describe below.


\begin{figure*}[t!]
	\centering
	\includegraphics[width=\textwidth]{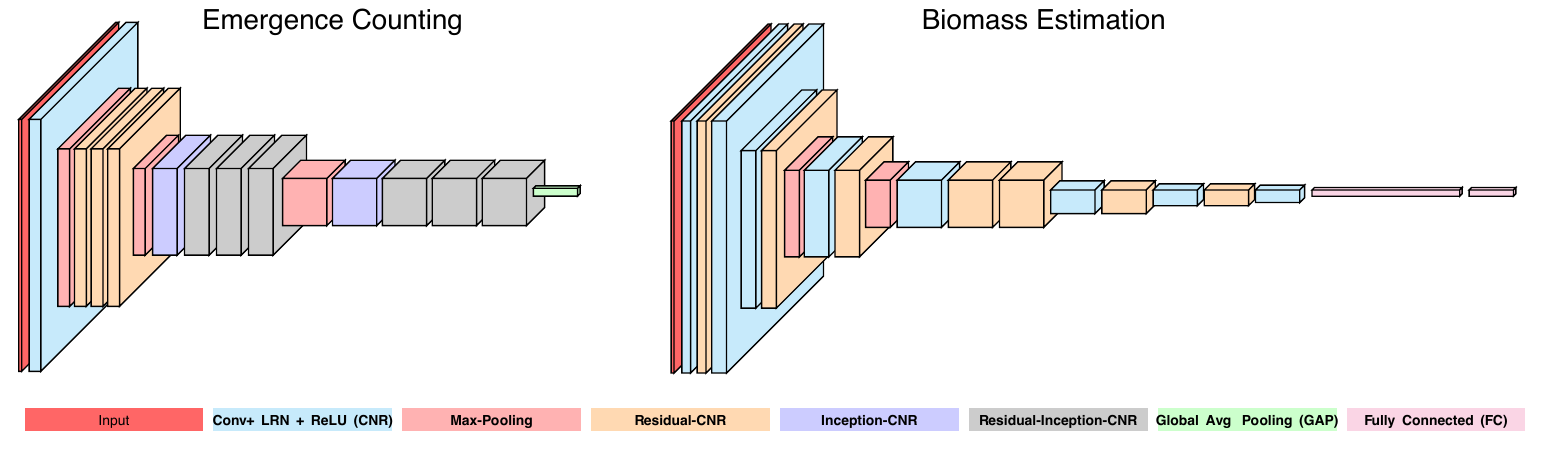}
    \caption{Emergence and biomass estimation architectures. We use $7\times 7$ receptive fields in the initial CNR block with unit stride. The number of filters after each max-pooling operation is doubled, except the first one for emergence counting. \textit{residual}-CNR is a simplified version of the residual block described in \cite{resnet}, where we keep the number of receptive fields constant inside the block. We use a simplified ``Inception" module \cite{inception-02}, where the number of input and output receptive fields are the same. Inside our Inception block, we employ half of the size of filters for $3\times3$ convolution, a quarter of the input size for the equivalent $5\times 5$ convolution, and half of the rest for pooling and unit convolution each. For the emergence network, to visualize the representations learned by our model, we use global average pooling (GAP) \cite{nin}.}
    \label{fig:arch_joint}
\end{figure*}

\subsubsection{Segmentation}

Our motivation for segmenting plot images into smaller patches is twofold. First, due to the very high resolution of plot images ($\sim2500\times7500$), it is not computationally feasible to do the emergence counting task on the whole image at once. Instead, either sequential or parallel counting over disjoint plant regions is required. Second, data-driven approaches, like deep learning, require many training samples, whereas we have only a few high-resolution plot images available for that purpose. Therefore, we generate non-overlapping patches of segmented plant regions to provide us with more than a hundred subsamples from each plot image for further training of the counting model.

From the design perspective, we relax the output of the segmentation module from exact segmentation to a soft or relaxed segmentation for several reasons. First, generating the exact ground-truth manually for images like the ones shown in Figure \ref{fig:gt} is a more tedious and time-consuming process than defining loose or relaxed contours around plants. Moreover, for deep networks, learning to count from the subsamples with exact vs. loose segmentations is similar since the background is uniform and so, it is unlikely that the model would pick up distinctive features from the background region. This claim is also validated by CAM \cite{cam-mit} visualizations of the network in the \textit{Experiments} section that show saliency in foreground regions. In addition, the wheat leaves are thin and partly occluded; therefore, going for precise segmentation could result in missing very thin or hard-to-detect regions of the plants which could deteriorate the counting performance since the model responsible for counting would assume the segregated leaves as different instances rather than a single one.

To perform soft segmentation with deep learning, we use the SegNet architecture \cite{segnet, aich-cvppp2017} rather than deconvolutional networks containing fully connected (FC) layers~\cite{deconvnet} with a many more training parameters. This is because 
 the problem we are dealing with is easier than the exact segmentation and much simpler than general multi-class semantic segmentation both in terms of the cardinality of the output categories and the nature of the domain since the diversity of the pixel intensities in a single plot image is highly restrained compared to that of natural images. Furthermore, our concern is not to get an overall-high precision segmentation mask, rather we are concerned with not missing plant regions in the image for the counting model afterward.

\subsubsection{Counting by Regression}

In this paper, we focus on different species of the crop wheat, which except the very late season, resembles mostly to grass crops. The leaves of such plants are the most deformable among all kinds of plants and crops, and so, a set of wheat plants in an image might appear in a combinatorially large number of variations. Thus, to successfully count the number of plants in the image, the deep model must be able to deal with such combinatorial number of deformations and resulting occlusions as much as possible.

As argued in the NIN paper \cite{nin}, a simple stack of convolutional layers with an over-complete set of filters followed by nonlinearity and pooling serve well when the underlying concepts to be learned via abstract representation are linearly separable. However, for highly nonlinear latent concepts, replacing plain convolutional blocks with small networks inside the basic architecture is already proved to be useful in several large-scale image classification tasks \cite{inception-01, inception-02}. Hence, we take inspiration from these works, where the representation in each layer is approximated from the dense multi-scale feature responses learned in the previous layer.  Also, we incorporate the concept of residual learning \cite{resnet} in our architecture, which we experimentally found to be useful for faster training in case of stacked-convolutional architecture for our task.

Therefore, in the design of our network as depicted in Figure \ref{fig:arch_joint}, four different convolutional blocks are used. Our initial convolutional block (CNR) is a simple convolution operation followed by local response normalization and rectified nonlinearity \cite{alexnet}. Next, we use a simplified residual version of the original residual block described in \cite{resnet}, in the sense that the number of feature maps is constant throughout the block from input to output. Also, for deeper layers, where the number of receptive fields is comparatively higher, we incorporate the ``Inception" version of CNR followed by the \textit{residual}-Inception version. All these modules are crafted to have the same input-output capacity.  Finally, for the ease of visualization of the salient regions detected by our model, we simply use the global average pooling (GAP) \cite{nin} layer. We experimented with different setups of fully connected layers instead of GAP and got slightly improved performance. However, we prefer visualization over those minor improvements to encourage further research based on visualization. Lastly, we have not used any pre-trained model because unlike classification problems, the capacity of the final layer does not scale up with the complexity of the counting task. In addition, opening up the full network for finetuning might result in significant overfitting due to comparatively smaller datasets.


\begin{figure}[]
	\centering
	\includegraphics[scale=0.30]{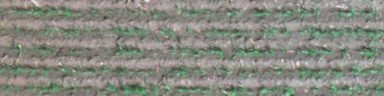}
	\includegraphics[scale=0.30]{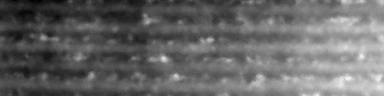} \\
	\includegraphics[scale=0.30]{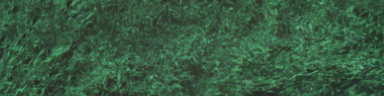}
	\includegraphics[scale=0.30]{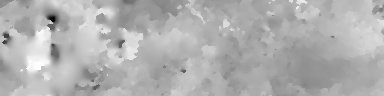} \\
	\includegraphics[scale=0.30]{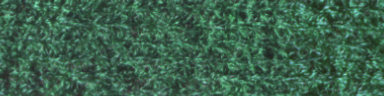}
	\includegraphics[scale=0.30]{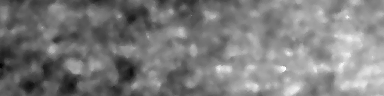}
    \caption{Sample RGB plot images (left) with corresponding DEMs (right) showing wheat plants from emergence as individual plants (top) to full crop canopy (middle) and during the reproductive stage (bottom). DEM values (height) converted to grayscale for visualization.}
    \label{fig:sample_RGBE}
\end{figure}

\subsection{Biomass Estimation}

For biomass estimation, we have both $5$ channel orthomosaics ($B$lue, $G$reen, $R$ed, $N$ear-infrared, red-$E$dge) and digital elevation maps (DEM). Sample RGB images are shown in Figure \ref{fig:sample_RGBE}.  The pixel values of the DEM files indicate the elevation of plants from the ground. Note that, the RGB images of the plots available for emergence counts in the previous section and biomass estimation here are from different sources. The plot images for biomass estimation are lower resolution ($\sim120\times480$) than those used for emergence counting (see Section~\ref{sec:datasets}).


above-ground biomass refers to the weight of all plant material above the ground. We expect that there is a relationship between biomass and height or elevation values of the $DEM$ images, but this relationship is difficult to observe from simple biomass versus elevation graphs. However, representing values from each plot as a different dimension in $\mathbb{R}^n$ space, we have found small angles ($[30\si{\degree}-32\si{\degree}]$ in our dataset) between the normalized elevation vector and the biomass vector. This suggests a nonlinear relationship between these two quantities and we take this as motivation for further computational analysis.

\begin{figure}[t]
	\centering
	\includegraphics[scale=0.30]{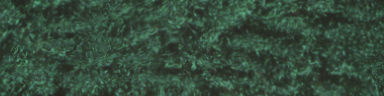}
	\includegraphics[scale=0.30]{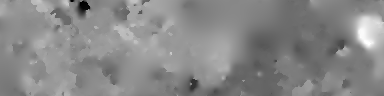} \\
	\includegraphics[scale=0.30]{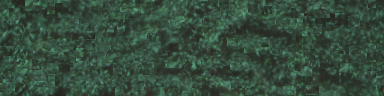}
	\includegraphics[scale=0.30]{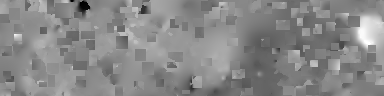} \\
	\includegraphics[scale=0.30]{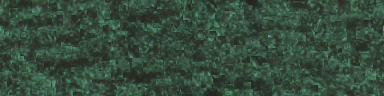}
	\includegraphics[scale=0.30]{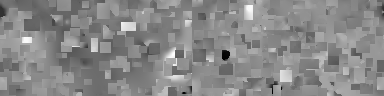} \\
	\includegraphics[scale=0.30]{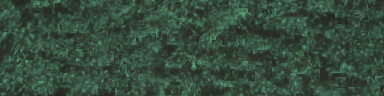}
	\includegraphics[scale=0.30]{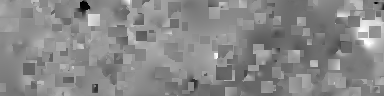} \\
	\includegraphics[scale=0.30]{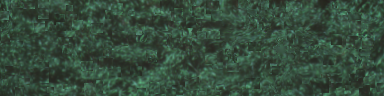}
	\includegraphics[scale=0.30]{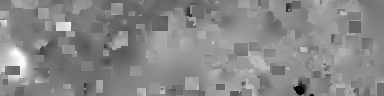}
    \caption{Sample RGB plot images (left) with corresponding DEMs (right) showing the original image (top row) and images generated by our \emph{RMRS} data augmentation procedure (other rows). DEM values (height) converted to grayscale for visualization.}
    \label{fig:sample_augment}
\end{figure}

Now, to apply any data-hungry models like deep learning to estimate biomass from these images, one of the main obstacles is the extremely low number of available samples ($\sim100$) for training and testing. One of the obvious ways to overcome this drawback is to figure out a suitable data-augmentation strategy. In this paper, we have devised a novel, simple and effective randomized data augmentation scheme that can be utilized to generate a sufficiently large number of augmented samples from each image. The idea is based on swapping similar superpixels in the image randomly. We call this approach the \textit{randomized minimal region swapping (RMRS)} algorithm. The steps of the \textit{RMRS} algorithm are as follows:

\begin{enumerate}
	\item Get the list of $K$ superpixels from RGB to gray-converted image and sort by their mean values.
	\item Generate a randomized list of length $N$ of the number of random swaps needed to generate the pool of $N$ augmented samples from a single image. The random integer values are in the range $[low, \floor{K/2}]$, where $low$ is the predefined threshold for the minimum number of swaps needed to create an augmented sample.
	\item For each number $r$ in the list generated in step 2, generate a randomized list of length $r$  of either even or odd superpixel indices in the range $[1, \floor{K/2}]$ and swap minimal rectangular regions between those even(odd) superpixels and their consecutive odd(even) counterparts in the sorted list. Even-odd consideration is necessary to avoid unaugmentation by repeated swaps.
\end{enumerate}

\begin{figure}[h!]
	\centering
	\includegraphics[scale=0.14]{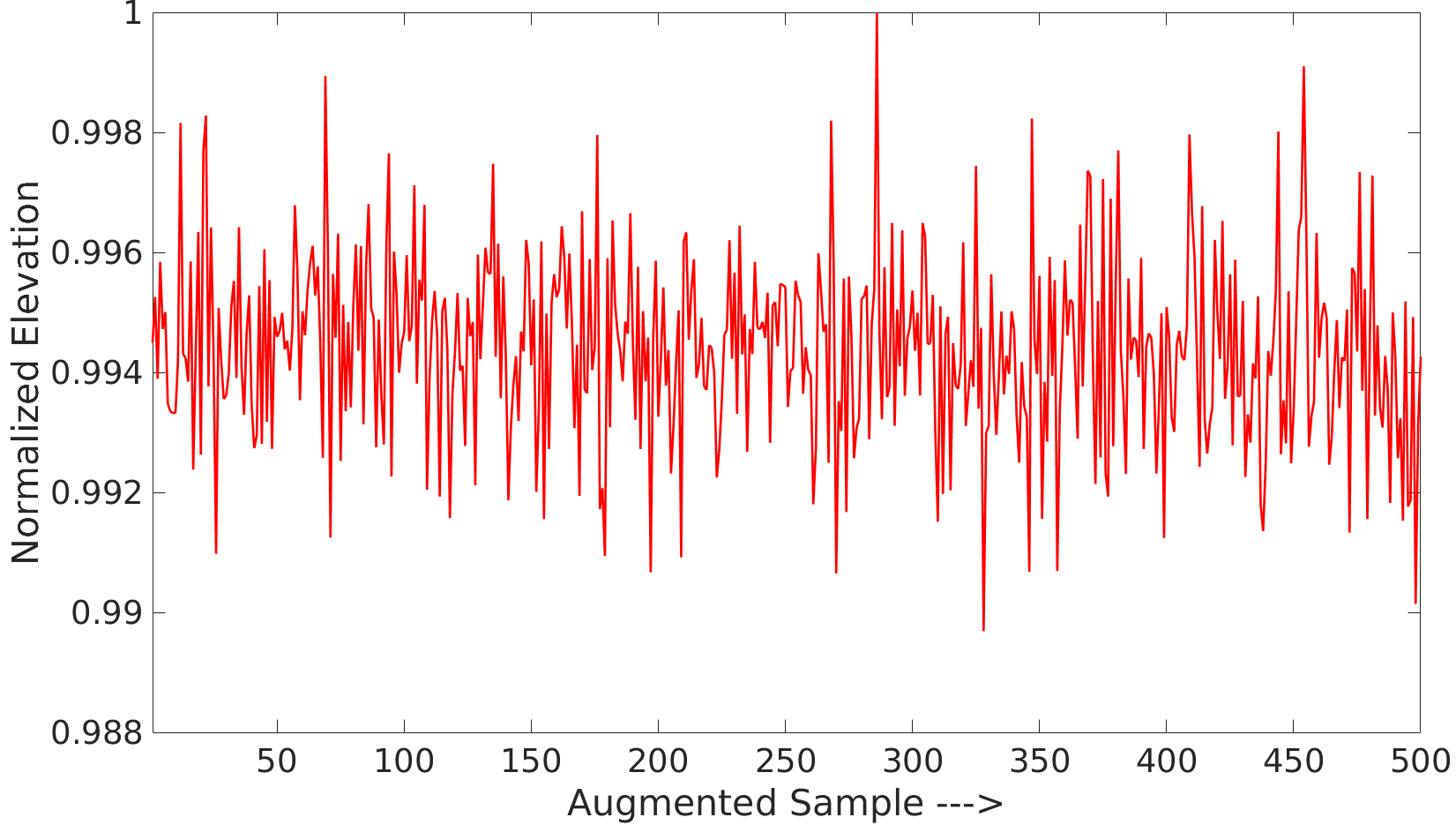}
    \caption{Normalized summation of the elevation for the samples augmented from a single image. The first point represents the elevation of the original sample and the rest $(499)$ are the augmented ones. The range of normalized elevation is in the range $[\sim0.99,1.0]$ indicating that the total elevation for all the samples are similar to the original.}
    \label{fig:elevation_plot}
\end{figure}

In our implementation, we use SLIC \cite{SLIC} as the superpixel algorithm. Figure \ref{fig:sample_augment} shows sample augmentation results for a single image along with the original one. As can be seen, it is impossible to identify the augmented samples as the artificial ones by looking only at RGB images, even though the corresponding DEMs appear to be highly discretized. Hence, as part of a further exploratory analysis, we plot the normalized summation of all pixel values or elevations of each DEM file for all the augmented samples along with the original one. Figure \ref{fig:elevation_plot} shows this normalized elevation plot for a single image and its augmented samples. As you can see, the normalized elevation varies in the range $[\sim0.99, 1.0]$, which means that although the augmented DEM files look different and discretized, the contents of the DEM pixels remain nearly constant after being augmented by the \textit{RMRS} algorithm.

In addition to increasing the number of training samples, augmenting data this way has another advantage as a byproduct. We hypothesize that the spatial relationships among the pixels in DEM images have little to do with the prediction of biomass since plants can be found in almost any region in the plot images. Therefore, the counting model should learn to map the pixel values from DEM images into the real-valued space of biomass in an almost spatially invariant manner. For data augmentation by \textit{RMRS} algorithm, new samples are just different permutations of the original one. From the practical standpoint, the interpretation might be that to generate an augmented sample, we swap the plants with similar color information within the plot. Thus, by learning to predict from this augmented dataset, the model may intrinsically learn a spatially invariant mapping from color and elevation to biomass.


Finally, we use a similar network architecture for biomass estimation (Figure \ref{fig:arch_joint}). The only difference between this model and the emergence count one is that the parameters and the placement of the computational blocks or layers are slightly modified to fit the model into this problem.

\section{Experiments}
\label{sec:experiments}

This section contains the experimental details of our work. First, we describe the datasets used for both tasks. Next, training procedure and implementational details of the networks are provided. Finally, the evaluation metrics are described and the evaluation results are reported in comparison to previous work along with the qualitative visualization of the salient regions.

\subsection{Datasets}
\label{sec:datasets}
The dataset used for emergence count consists of 274 wheat \textit{(Triticum durum)} plots of $1.5m\times3.7m$ area. High-resolution aerial images ($\sim2500\times7500$ pixels per plot) were captured for each plot by walking through the field with a GoPro Hero 5 camera \cite{gopro} mounted on a monopod with a gimbal for stabilization. Covering plots with this device has the advantage of getting very high-resolution images appropriate for detailed computational analysis compared to other remote sensing technologies.

For biomass estimation, we used aerial drone images for 48 wheat (Triticum aestivum) plots for two dates: June 27 and July 20, 2016. The UAV images have been captured using a MicaSense RedEdge camera \cite{micasense} on a DraganFly Commander drone \cite{draganfly}. The RedEdge camera includes five different sensors, one for each band: $Blue (\sim 465-485 nm)$, $Green (\sim 550-570 nm)$, $Red (\sim 658-678 nm)$, $NIR (\sim 820-860 nm)$, and $Red Edge (\sim 707-727 nm)$. The output from these sensors was post-processed using the Agisoft Photoscan \cite{agisoft} to generate an orthomosaic image and digital elevation map. For each of these dates, manual ground truth measurement of biomass have also been conducted. For manual counting, plants were cut randomly from the plots at ground-level using sickles, dried, and then weights of those plants were noted. The dataset is randomly split into two equal subsets for training and testing.

\subsection{Training and Implementation}

We used Torch \cite{torch} as the deep learning framework. To train the segmentation network, we generated $0.25M$ sub-samples of size $224\times224$ from 10 high-resolution plot images. The network was trained for $30$ epochs over this augmented dataset. SGD-momentum was used as the optimizer with a fixed learning rate, momentum, and weight decay of $0.01, 0.9,$ and $0.0001$ respectively, over the training period.


Both the emergence count and biomass estimation networks were trained with similar parameter settings. Adam optimizer \cite{adam} was used with learning rate and weight decay both set to $0.0001$. Absolute value and Smooth L1 measures \cite{fast-rcnn} are used as the error criteria (loss functions) for training emergence and biomass models, respectively. For emergence network training, we slowed down the training rate later based on our observation of the training statistics. Training for the emergence network was conducted for $100$ epochs, whereas the biomass estimation network was trained with different combinations of input channels for $50$ epochs with the same initial parameter settings. We will provide the link for pre-trained models and codes in the final version of this paper.

Note that the emergence count network was trained on $7855$ patches extracted from $37$ images and their slightly augmented versions. On the other hand, the biomass network was trained with about $0.15M$ augmented training samples generated by the \textit{RMRS} algorithm from $48$ plot samples. Codes, pre-trained models, and datasets are publicly available here. \footnote{\texttt{https://github.com/p2irc/deepwheat\_WACV{-}2018}}

\subsection{Evaluation}

Here, we provide three evaluations of our approach. First, we assess the performance of our segmentation network for generating relaxed binary segmentations. Next, both emergence count and biomass estimation networks are evaluated based on the metrics listed in Equation \ref{eq:metrics_evaluation} below. Among these metrics, we take \textit{MAD} and \textit{SDAD} from the leaf counting benchmark \cite{aich-cvppp2017}. The other is simply a variant of these measures. In addition, we provide CAM visualization for the emergence counting model.

\begin{table}[!htbp]
\centering
{\renewcommand{\arraystretch}{1.5}
\begin{tabular}{|l|l|}
\hline
$\text{Precision} = \frac{\text{True Positive}}{\text{True Positive} \, + \, \text{False Positive} }$ & 85.59 \\ \hline
$\text{Recall} = \frac{\text{True Positive}}{\text{True Positive} \, + \, \text{False Negative}}$ & 83.76 \\ \hline
$\text{Accuracy} = \frac{\text{True Positive} \, + \, \text{True Negative}}{\textit{All}}$ & 93.76 \\ \hline
\end{tabular} }
\caption{Binary segmentation results}
\label{tab:binseg}
\end{table}

\textbf{Emergence evaluation: } Precision, recall, and accuracy are measured to evaluate the segmentation network (Table \ref{tab:binseg}). Results for precision ($\sim86\%$) and recall ($\sim84\%$) are a somewhat low because the ground truth segmentations are not precise, but loosely defined contours covering all the plant regions in the images. To justify our outputs, we have visually checked almost all the test segmentation results and find almost no plant regions undetected by the network.

\begin{equation}
\begin{cases}
a_i, t_i = \text{actual and target counts for } i^{th} sample \\
N = \text{Number of samples} \\
\textit{\%Difference(\%D)} = \frac{\sum_{i}|a_i-t_i|I_{[a_i-t_i \neq 0]}}{\sum_{i}t_i} \\
\textit{Mean Absolute Difference (MAD)} = \frac{\sum_{i}|a_i-t_i|}{N} \\
\textit{Std Absolute Difference (SDAD)} = \sqrt{\frac{\sum_{i}(|a_i-t_i|-MAD)^{2}}{N-1} }
\end{cases}
\label{eq:metrics_evaluation}
\end{equation}

\begin{table}[h!]
\centering
\begin{adjustbox}{scale=0.95}
\begin{tabular}{|
>{\columncolor[HTML]{FFFFFF}}c |
>{\columncolor[HTML]{FFFFFF}}c |
>{\columncolor[HTML]{FFFFFF}}c |
>{\columncolor[HTML]{FFFFFF}}c |}
\hline
Problem & MAD & SDAD & \%D \\ \hline
Prev. Leaf Counting~\cite{aich-cvppp2017} & 1.62 & 2.30 & - \\ \hline
Plain Architecture & 1.13 & 1.42 & 27.04 \\ \hline
Inception Architecture & 1.08 & 1.38 & 25.78 \\ \hline
Our Emergence Counting & \textbf{1.05} & \textbf{1.40} & \textbf{25.08} \\ \hline
\end{tabular}
\end{adjustbox}
\captionsetup{justification=centering}
\caption{Evaluation metrics for the emergence count model}
\label{tab:emergence}
\end{table}


Table \ref{tab:emergence} lists the evaluation metrics for our emergence counting network. As stated in the introduction, we did not find appropriate literature to benchmark our approach. The closest approach is the one used for Arabidopsis and Tobacco leaf counting problem \cite{aich-cvppp2017}. 
We achieve \textit{\%D} of 25\% and \textit{MAD} and \textit{SDAD} of $1.05$ and $1.40$ which is more accurate than previously reported results for one of the best leaf counting system currently available. These results are notable because counting wheat plants with thin, overlapping leaves from outdoor images is substantially more difficult than counting leaves from indoor images of rosette plants (as discussed in the Introduction and illustrated in Figure~\ref{fig:emergence_difficult}). We have also included the results for the corresponding plain and Inception-only version to justify the additional complexity of our final model. The plain network was trained for twice the number of epochs than others.

\begin{figure}[]
	\centering
	\includegraphics[scale=0.34]{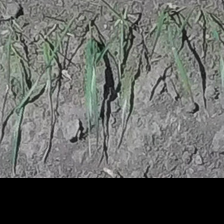}
	\includegraphics[scale=0.34]{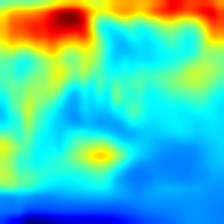}
	\includegraphics[scale=0.34]{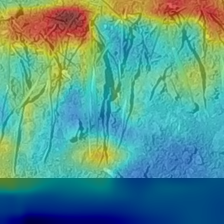} \\
	\includegraphics[scale=0.34]{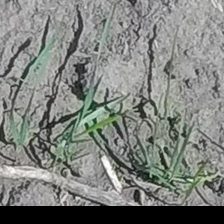}
	\includegraphics[scale=0.34]{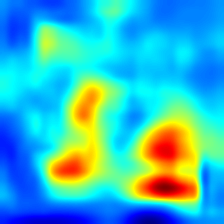}
	\includegraphics[scale=0.34]{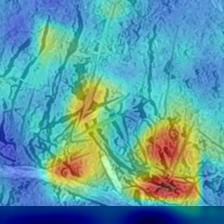}
    \caption{Sample RGB images (left), their CAM \cite{cam-mit} visualizations (middle), and superimposed images (right). Note that, RGB images are padded by black to maintain a constant size of $224\times 224$. Red and blue indicate the most and the least significant regions responsible for emergence counting. As you can see, the plant bases are detected as the most salient regions (red) which the experts also use for counting followed by the leaves (yellow).}
    \label{fig:cam}
\end{figure}

The salient regions detected by our counting model for sample RGB images are shown as heatmaps, generated by CAM \cite{cam-mit}, in Figure \ref{fig:cam}. Although in the original paper, CAM is used to visualize class-specific mapping of the salient regions, for our counting task, it can also be used for visualizing the regions responsible for making the counts. As already discussed, the bases of leaf-clusters are the most significant parts for successful counts followed by dense regions of overlapping leaves. The sample heatmaps also follow this counting strategy. In the heatmaps, the bases of the plants are marked with red (highest saliency) followed by the leaves with yellow, which clearly indicates that our model is capable of identifying the correct regions in the images responsible for counting. Nonetheless, our percentage deviation is a bit high because of the inherent difficulty of counting the plants due to severe occlusion and large leaf deformations.
To enable CAM visualization, we cut out additional fully connected layers, which had provided a slight performance boost, but the resulting visualization provides more valuable insight into the learning process for plant counting.


\textbf{Biomass evaluation: } Table \ref{tab:comp_biomass} contains the same metrics as in Equation \ref{eq:metrics_evaluation} for biomass models trained with different input channel combinations. Here, $H$, $R$, $G$, $B$, $N$, and $E$ stand for $DEM$, $Red$, $Green$, $Blue$, $NIR$, and $Red Edge$ channels, respectively.

As can be seen, the model trained with only $H(DEM)$ as input gives \textit{\%D} of $\sim26\%$, which is $\sim4\%$ and $\sim2\%$ lower than the model trained with $RGBH$ and all the channels. At this point, it is unclear whether the deep learning model takes care of any of the RGB texture in the biomass image. Intuitively, although color information or greenness of the RGB image might be important, the texture information is not that significant for biomass estimation. However, there is a high variance in the color information under different weather conditions. For instance, if the weather is overcast, crops will appear dark-green, for sunny weather, it will be yellowish-green, and so on. Another critical issue is that after augmenting data using \textit{RMRS} algorithm, albeit the very local texture property and the total energy of the images are more or less preserved, semi-local texture property is destroyed. We are not sure whether this lack of semi-local texture causes the network trained with RGBE input to perform poorer than the one with only DEM input. This issue can only be explored further if sufficient raw training samples are available in future.

On the other hand, the fact that the model works better when two extra non-visible wavelengths, such as, $NIR$ and $Red Edge$, are provided along with $RGB$, is consistent with the plant science literature \cite{sellers1985} where vegetation indices extracted from hyperspectral and visible wavelength data are used as strong indicators of photosynthetic measurements of plants. However, the utility of hyperspectral data for biomass estimation is still an open question.


\begin{table}[h!]
\centering
\begin{adjustbox}{scale=0.87}
\begin{tabular}{|
>{\columncolor[HTML]{FFFFFF}}c |
>{\columncolor[HTML]{FFFFFF}}c |
>{\columncolor[HTML]{FFFFFF}}c |
>{\columncolor[HTML]{FFFFFF}}c |
>{\columncolor[HTML]{FFFFFF}}c |
>{\columncolor[HTML]{FFFFFF}}c |}
\hline
Method & MAD & SDAD & \%D \\ \hline
$H_1$+MARS\cite{laurin2016} & 1.66 & 2.03 & 29.61 \\ \hline
$H_2$+PLS\cite{laurin2014} & 3.86 & 2.72 & 68.92 \\ \hline
$H_2$+MARS\cite{laurin2016, laurin2014} & 1.74 & 2.07 & 30.96 \\ \hline
$OH_3$+MLR\cite{bendig2015} & 1.67 & 1.63 & 29.67 \\ \hline
Ours ($RGBH $) & 1.67 & 2.05 & 29.75 \\ \hline
Ours ($RGBNEH$) & 1.53 & \textbf{1.62} & 27.38 \\ \hline
Ours ($H$) & \textbf{1.45} & 2.05 & \textbf{25.88} \\ \hline
\end{tabular}
\end{adjustbox}
\caption{Comparison of biomass estimation metrics to other methods and with different input channels ($H \equiv DEM$, $R$ed, $G$reen, $B$lue, $N$ear-infrared, and red$E$dge)}
\label{tab:comp_biomass}
\end{table}

In Table \ref{tab:comp_biomass}, we provide a comparison against the recent literature. We implemented the methods described in \cite{laurin2016, laurin2014, bendig2015} on our data for comparison. These three papers reported the effect of different feature combinations from the set of simple statistical features based on height and different vegetation indices as the predictor variables for their regression models. In this table, we use the combination of features that performed best on our dataset. $H_1$, $H_2$, and $H_3$ indicates slightly different variations statistical height features and $OH_3$ stands for the combination of $H_3$ and \textit{Optimized Soil-Adjusted Vegetation Index (OSAVI)}. Also, \textit{MARS (Multivariate Adaptive Regression Splines)}, \textit{PLS (Partial Least Squares)}, and \textit{MLR (Multivariate Linear Regression)} are different linear and nonlinear regression algorithms. As can be seen, even with such tiny amount of original training data, the best performance of our deep model (trained with $DEM(H)$) is $\sim4\%$ better than the recent nonlinear regression model for biomass.

\section*{Acknowledgment}
This research was undertaken thanks in part to funding from the Canada First Research Excellence Fund and the Natural Sciences and Engineering Research Council (NSERC) of Canada. We also thank Seungbum Ryu and the USask field crew for providing biomass data.

\section{Conclusion and Future Work}
\label{sec:conclusion}
In this paper, we have developed three different deep learning models for segmenting plant regions, counting plants, and estimating biomass from aerial field images. Our results show better biomass estimation accuracy than previous methods and better accuracy for outdoor emergence counting as compared to previous studies of indoor leaf counting. Although we have only evaluated our model on particular species of wheat, we expect that our design methodology allows for generalization of these models to other types of crops with minimal changes. As future work, we plan to evaluate our networks with other crops that have different plant morphologies, such as pulses and oilseeds. We also plan to further investigate if estimation accuracy for these phenotypic traits can be improved with larger datasets in subsequent growing seasons, as well as the use of digital elevation maps together with non-visible wavelengths of light as input for biomass estimation.

{\small
\bibliographystyle{ieee}
\bibliography{egbib,agriculture}
}

\balance

\newpage

\end{document}